\UseRawInputEncoding
\documentclass[runningheads]{llncs}
\usepackage{orcidlink}
\usepackage{multirow}
\usepackage{tabularx}
\usepackage{amsmath} 
\usepackage{color}
\usepackage{comment}
\usepackage{subcaption}

\usepackage[utf8]{inputenc} 
\usepackage[T1]{fontenc}

\usepackage[T1]{fontenc}
\usepackage{graphicx}

\usepackage{color}

\begin{document}
\title{DADO: A Depth-Attention framework \\ for Object Discovery}
%
%

\authorrunning{Anonymous}

\author{
  Federico Gonzalez\inst{1,2} \orcidlink{0000-0002-3993-4084} \and 
  Estefania Talavera \inst{4} \orcidlink{0000-0001-5918-8990} \and 
  Petia Radeva\inst{1,3} \orcidlink{0000-0003-0047-5172}
}

\authorrunning{F. Gonzalez et al.}

\institute{
  Universitat de Barcelona, Gran Via de les Corts Catalanes, 585, Barcelona, 08007, Spain \\
  \email{petia.ivanova@ub.edu}
  \and
  Universidad Nacional de Tierra del Fuego, Fuegia Basket 251, Ushuaia, 9410, Argentina \\
  \email{fgonzalez@untdf.edu.ar}
  \and
  Institut de Neurosciències, University of Barcelona, Passeig de la Vall d’Hebron, 171, Barcelona, 08035, Spain \\
  \and
  University of Twente, Drienerlolaan 5, 7522 NB, Enschede, The Netherlands \\
  \email{e.talaveramartinez@utwente.nl}
}

\maketitle

\begin{abstract}
 Unsupervised object discovery, the task of identifying and localizing objects in images without human-annotated labels, remains a significant challenge and a growing focus in computer vision. In this work, we introduce a novel model, DADO (Depth-Attention self-supervised technique for Discovering unseen Objects), which combines an attention mechanism and a depth model to identify potential objects in images. To address challenges such as noisy attention maps or complex scenes with varying depth planes, DADO employs dynamic weighting to adaptively emphasize attention or depth features based on the global characteristics of each image. We evaluated DADO on standard benchmarks, where it outperforms state-of-the-art methods in object discovery accuracy and robustness without the need for fine-tuning.

\end{abstract}

\section{Introduction}
\label{sec:introduction}

Object discovery is the unsupervised task of identifying and localizing objects in images or videos without prior knowledge of their categories or reliance on labeled data, distinguishing it from supervised object detection \cite{lee_learning_2011,sivic_discovering_2005}. The field's terminology is evolving, often segmenting the process into stages: unsupervised saliency to separate foreground, single-object discovery to localize a primary object, multi-object discovery (or zero-shot unsupervised object detection) to identify multiple significant objects, and class-agnostic instance segmentation for individual object masks \cite{simeoni_unsupervised_2023-1,villa-vasquez_unsupervised_2024}.

While large Visual Language Models (VLMs) like CLIP \cite{radford2021learningtransferablevisualmodels} or SAM \cite{kirillov2023segany} exhibit impressive capabilities in vision-language tasks, such as zero-shot classification and retrieval through associating visual content with semantic descriptions, they are not inherently equipped for unsupervised object discovery of novel, unseen objects. Their abilities are tied to concepts seen during training, and lack autonomous spatial localization or object proposal generation mechanisms necessary for true unsupervised discovery without further adaptation.

Object discovery remains crucial for understanding visual data in a category-agnostic, cost-efficient manner, particularly for novel object detection or in data-scarce environments where labeled data is unavailable. It complements large pre-trained models by enabling broader generalization and adaptation in unconstrained, real-world settings.

Self-supervised learning (SSL) is fundamental to modern object discovery, allowing models to learn robust visual representations directly from unlabeled data, providing flexibility for dynamic environments. Foundational SSL methods like SwAV \cite{caron_unsupervised_2021}, SimCLR \cite{chen_simple_2020}, BYOL \cite{grill_bootstrap_2020}, MAE \cite{he_masked_2022}, and MoCo \cite{he_momentum_2020} were instrumental, with DINO \cite{caron_emerging_2021,oquab_dinov2_2023} becoming a key method for extracting high-quality unsupervised semantic features.

A central challenge in the unsupervised paradigm is inferring what constitutes an "object" and its relative importance without labels, avoiding potentially limiting assumptions about its visual characteristics, location, or size. Although classical definitions often describe objects as salient structures with distinct, separable boundaries \cite{alexe_what_2010,biederman_recognition-by-components_1987,marr_vision_2010}, rigid adherence to such assumptions can hinder the design of versatile discovery systems.

To address these inherent challenges, we propose DADO, a novel unsupervised object discovery approach. DADO integrates attention and depth cues with a dynamic entropy-balancing mechanism for image-feature-based weighting, thereby improving accuracy and robustness. Our comprehensive evaluation on standard benchmarks shows that DADO consistently outperforms state-of-the-art object discovery methods.

\section{Related work}
\label{sec:related}

Unsupervised object discovery, identifying objects without appearance-based assumptions, remains an open challenge. Before the SSL era, methods such as LOD \cite{vo2021largescaleunsupervisedobjectdiscovery} and rOSD \cite{vo2020unsupervisedmultiobjectdiscoverylargescale} achieved notable results in multi-object discovery. The widespread availability of SSL, Vision Transformers (ViT), and large pre-trained models, particularly DINO, has significantly advanced this field, with DINO features forming the basis for many recent approaches.

Prominent SSL-based methods include LOST \cite{simeoni_localizing_2021}, which leverages attention maps from pre-trained ViTs to identify a 'seed' patch and derive object masks and bounding boxes. TokenCut \cite{wang2023tokencutsegmentingobjectsimages} constructs a graph from self-attention tokens and employs normalized graph cuts by spectral clustering to delineate foreground objects. However, both LOST and TokenCut often focus on the single most salient region, grappling with images containing multiple or overlapping objects.

Addressing this limitation, MOST \cite{rambhatla_most_2023} also uses DINO features, but applies fractal analysis through box-counting on patch similarities to identify and cluster multiple foreground object 'pools'. In contrast, FOUND takes an inverse approach by using DINO features to discover the background, considering everything else as the foreground.

In addition to appearance-based methods, self-supervised depth estimation has proven valuable for object localization \cite{chen2019learningsemanticsegmentationsynthetic,hoyer2021improvingsemisuperviseddomainadaptivesemantic,safadoust2023multiobjectdiscoverylowdimensionalobject}. This utility comes from the observation that depth discontinuities often align with object boundaries \cite{hoyer2021waysimprovesemanticsegmentation,Vandenhende_2021}, facilitating the separation of foreground and background elements and helping in handling occlusion.

\section{The DADO Method}
\label{sec:method}

\begin{figure*}[h]
    \centering
    \includegraphics[width=\textwidth]{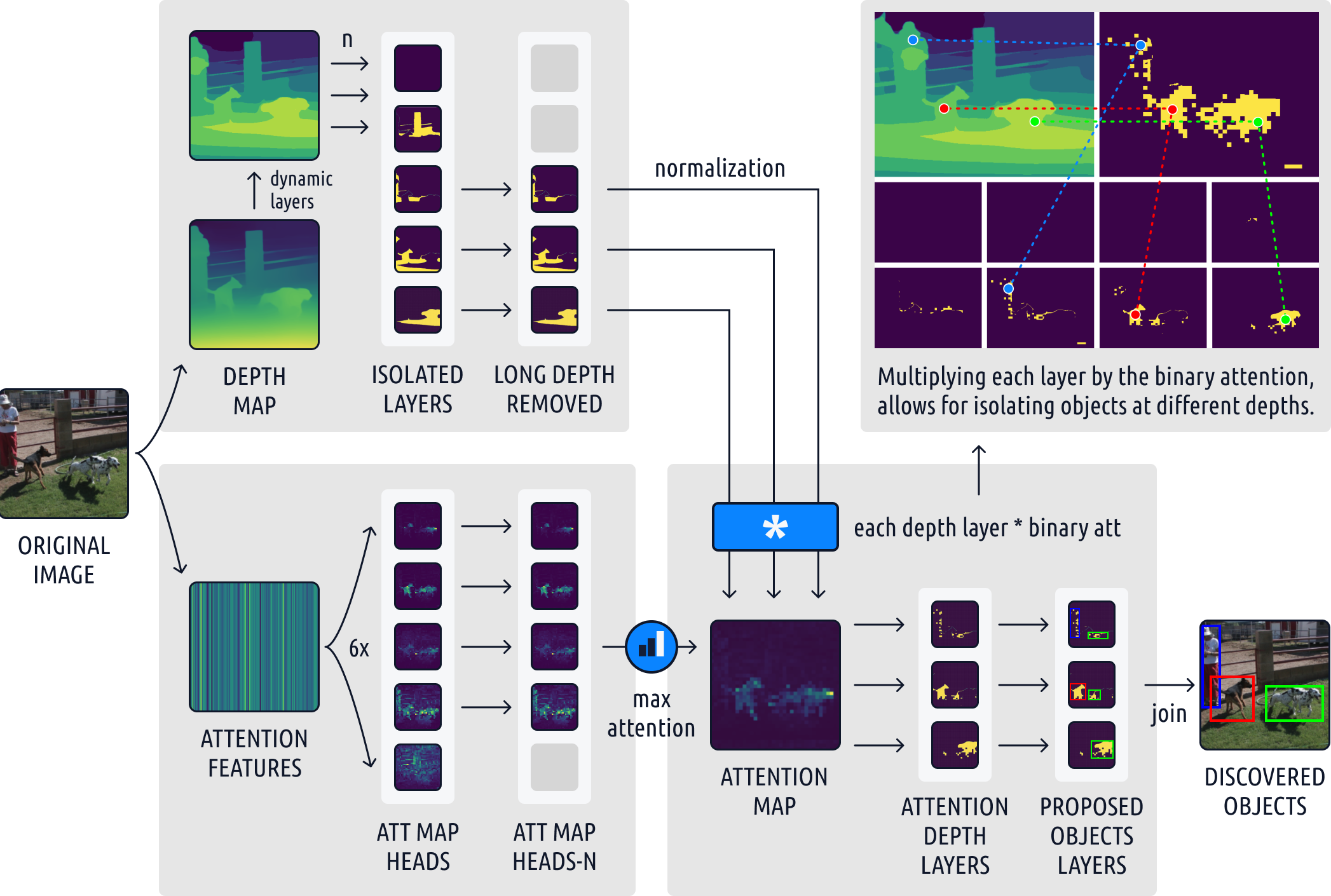}
    \caption{Our proposed DADO framework}
    \label{fig:pipeline}
    \vspace*{-.8cm}
\end{figure*}

DADO leverages SSL to extract attention insights and depth estimation techniques to capture structural features of the scene (see Figure \ref{fig:pipeline}). Our pipeline takes a single RGB input image and computes both depth estimation and attention features. It then segments the scene into discrete depth layers and, in parallel, constructs a global attention map. By combining each depth layer with the attention map, DADO isolates candidate objects at varying depth ranges. The code is publicly available at link \footnote{https://github.com/fedegonzal/dado}.

\paragraph{\textbf{Depth Map Estimation}}

We extract the depth maps using a Dense Prediction Transformer (DPT) model \cite{ranftl_vision_2021}. This model produces a dense depth representation of the scene, capturing its structural characteristics and spatial configuration. The depth map distinguishes between foreground and background elements, providing essential cues for object separation based on depth segmentation.

After generating the depth map, we process it to isolate distinct depth layers. This step is crucial for separating objects located at different depths, particularly when one object is positioned in front of another. To extract these isolated layers, we generate adaptive depth intervals from the depth map by analyzing the distribution of depth values using a histogram. Significant peaks are identified on the basis of  their prominence, under the assumption that they correspond to dominant depth layers in the scene. Around each peak, a depth bin is defined with a configurable overlap (20\% by default), resulting in a set of meaningful, data-driven depth ranges that reflect the underlying structure of the environment. This process is represented in Figure~\ref{fig:pipeline} as \textit{dynamic layers}.

We then generate $n$ binary images, each corresponding to one of these depth ranges. In each binary image, pixels within the corresponding depth bin are set to ones, while all others are set to zero. Since the farthest layers typically correspond to background regions, we discard those layers. The result is a stack of binary images representing the foreground information at different depth levels.

\paragraph{\textbf{Attention Map Generation}}

Concurrently, we extract attention maps using DINO ViT Small. The attention mechanism within the ViT enables the model to focus on distinct and relevant image features by generating weight maps that highlight these regions. DINO produces six CLS attention heads, which we aggregate by taking the maximum value across heads for each pixel. The resulting attention map provides a coarse visual understanding of which regions in the image are of interest.

\[
\texttt{attMask} = \max(\texttt{attention\_map\_heads}) .
\]

\paragraph{\textbf{Integrating Attention and Depth Maps}}

DADO integrates attention and depth information. Before combining attention and depth layers, \textbf{we normalize} each one to the range [0, 1] to produce binary masks. This normalization ensures compatibility between both types of data, facilitating a meaningful combination.

Object-centric images typically exhibit well-defined attention maps where the model clearly focuses on the prominent object(s) \cite{naseer2021intriguingpropertiesvisiontransformers,trivedy2025learningobjectfocusedattention}. In contrast, complex images containing multiple or sparsely distributed objects can result in more diffuse or "noisy" attention maps \cite{mehrani2023selfattentionvisiontransformersperforms,raghu2022visiontransformerslikeconvolutional}, where the focus is less concentrated and may be scattered across various regions. Leveraging the image's entropy, we apply dynamically adjusted weights to combine both depth and attention maps, ensuring that the contribution of each model is balanced according to the structural complexity of the scene.

\paragraph{\textbf{Depth-Attention-Driven Object Isolation}}

Given the pipeline described above, we define DADO as the combination of depth and attention information, weighted dynamically according to the complexity of the input image.

Let \texttt{attMask} be the global attention map, and let \texttt{depthLayer\_i} represent the $i$-th foreground depth layer, obtained by histogram-based segmentation of the depth map. For each layer, we compute a weighted combination with the attention map as follows:

\begin{equation}
\text{DADO}_i = w_a \cdot \texttt{attMask}  \times w_d \cdot \texttt{depthLayer}_i,  
\label{eq.1}
\end{equation}
\noindent where $w_a$ and $w_d$ are dynamically determined based on the entropy or structural complexity of the input image.

The cross-correlation $CC$ between the attention map and the depth map is calculated as the mean of their element-wise product:
\[
CC = \frac{1}{N} \sum_{i=1}^{N} \left( \texttt{atts\_normalized}[i] \times \texttt{depth\_normalized}[i] \right),
\]
where \( N \) is the number of elements in the maps.
If $CC$ exceeds a threshold (e.g., 0.5), both weights are set to 0.5, indicating that both maps are reliable:
\[
\text{if } \texttt{CC} > 0.5, \quad w_a = w_d = 0.5.
\]

Otherwise, the weight $w_a$ is calculated based on the attention sparsity:
\[
w_a = \frac{1}{1 + \texttt{attention\_sparsity}},
\]
where the lower attention sparsity (more concentrated attention) results in a higher weight.
The weight $w_d$ is determined by the depth gradient consistency:
\[
w_d = \texttt{depth\_gradient\_consistency}.
\]

Following Equation (\ref{eq.1}), \textbf{DADO} is the set of all combined maps for each depth layer:
\begin{equation}
\textbf{DADO} = \left\{ \text{DADO}_i \;\middle|\; i = 1, \dots, n \right\}.
\end{equation}

This formulation allows the model to emphasize either attention or depth features, adapting to the characteristics of each scene, and improving object isolation across different depth levels.

\paragraph{\textbf{Adaptive Thresholding}}

To threshold the attention-depth map ($att\_depth$), we compute the average of the mean and standard deviation of the attention-depth values:
\[
\tau = \frac{\text{mean}(\texttt{att\_depth}) + \text{std}(\texttt{att\_depth})}{2}.
\]

This threshold $\tau$ effectively sets a central value between the typical range and the variability of the map. Next, the attention-depth map is thresholded based on this value. 
Any element in the map with a value less than or equal to the threshold is set to 0, effectively removing or "masking" that part of the map. On the other hand, values greater than the threshold are set to 255, marking these parts of the map as the regions of interest or relevant features. The attention depth map is then binarized as follows:

\[
\mathtt{att\_depth\_masked}(i,j) =
\begin{cases}
0, & \text{if } \mathtt{att\_depth\_masked}(i,j) \le \tau \\
255, & \text{if } \mathtt{att\_depth\_masked}(i,j) > \tau
\end{cases}
\]

This process results in a binary mask, where values above $\tau $ are highlighted as significant, while the rest of the map is suppressed. This thresholding technique helps to emphasize the most significant parts of the attention-depth map, based on its statistical properties (mean and standard deviation). $\tau $ is set at a midpoint between the mean and standard deviation, and the values above $\tau $ are considered important for further processing, while the values below it are masked out.

\paragraph{\textbf{Bounding Box Estimation}}

From the thresholded composite map, potential object regions are delineated using morphological operations and contour detection methodologies. This dual-step approach facilitates the generation of robust bounding boxes, which are refined using Soft Non-Maximum Suppression to minimize overlap and optimize object localization.

\section{Validation}
\label{sec:experiments}

\paragraph{\textbf{Datasets}} 
To facilitate a comparable evaluation, our work uses Pascal VOC 2007 (VOC07) \cite{pascal-voc-2007}, and VOC 2012 (VOC12) \cite{pascal-voc-2012}.

\paragraph{\textbf{Evaluation metrics}} 
We evaluate our unsupervised object discovery method using two complementary metrics: Correct Localization (CorLoc) \cite{Deselaers_2010_Localizing_Objects_While_Learning_Their_Appearance} and object discovery Average Precision (odAP) \cite{cho2015unsupervisedobjectdiscoverylocalization}. CorLoc measures the percentage of images where the dominant object is correctly localized, based on an Intersection over Union (IoU) threshold of 0.5. It only assesses whether at least one object is detected per image, without considering multiple object retrieval. To address this limitation, we also report results using odAP, which extends standard Average Precision to the unsupervised setting. odAP evaluates both the accuracy of object localization and the ability to retrieve all relevant objects, summarizing performance via precision-recall curves. We compare our method against state-of-the-art approaches. We benchmark against MOST \cite{rambhatla_most_2023}, LOST \cite{simeoni_localizing_2021}, and TokenCut \cite{wang2023tokencutsegmentingobjectsimages} using CorLoc. For odAP, we compare with MOST \cite{rambhatla_most_2023}, rOSD \cite{vo2020unsupervisedmultiobjectdiscoverylargescale}, and LOD \cite{vo2021largescaleunsupervisedobjectdiscovery}. 

\begin{table*}[ht]
\centering
\begin{minipage}[t]{0.48\textwidth}
\centering
\caption{Object Discovery Average Precision (odAP) evaluation on Multiple Object Localization.}
\label{tab:odAP}
\begin{tabularx}{\textwidth}{X|l|r|r}
\hline
\textbf{Method} & Year & \textbf{VOC07} & \textbf{VOC12} \\
\hline
rOSD \cite{vo2020unsupervisedmultiobjectdiscoverylargescale} & 2020 & 4.3 & 5.27 \\
LOD \cite{vo2021largescaleunsupervisedobjectdiscovery} & 2021 & 4.5 & 5.34 \\
MOST \cite{rambhatla_most_2023} & 2023 & \textbf{6.4} & -- \\
DADO (ours) & 2025 & 6.2 & \textbf{5.9} \\
\hline
\end{tabularx}
\end{minipage}
\hfill
\begin{minipage}[t]{0.48\textwidth}
\centering
\caption{CorLOC performance comparison of DADO with Related Work.}
\label{tab:CorLOC}
\begin{tabularx}{\textwidth}{l|l|r|r}
\hline
\textbf{Method} & \textbf{Year} & \textbf{VOC07} & \textbf{VOC12} \\
\hline
rOSD \cite{vo2020unsupervisedmultiobjectdiscoverylargescale} & 2020 & 54.5 & 55.3 \\
LOD \cite{vo2021largescaleunsupervisedobjectdiscovery} & 2021 & 53.6 & 55.1 \\
LOST \cite{simeoni_localizing_2021} & 2021 & 61.9 & 64.0 \\
TokenCut \cite{wang2023tokencutsegmentingobjectsimages} & 2023 & 68.8 & 72.1 \\
MOST \cite{rambhatla_most_2023} & 2023 & 74.8 & \textbf{77.4} \\
DADO (ours) & 2025 & \textbf{78.3} & 74.2 \\
\hline
\end{tabularx}
\end{minipage}
\vspace*{-5mm}
\end{table*}

\paragraph{\textbf{Results}}
\label{sec:results}
DADO achieves competitive results in both single-object discovery and multi-object discovery. Table~\ref{tab:odAP} shows the performance of our method with an odAP of 6.2 for VOC07 and 5.9 for VOC12. For CorLOC, DADO achieves 78.3\% on VOC07 and 74.2\% on VOC12, as shown in Table~\ref{tab:CorLOC}.

Large pre-trained models like DINO have significantly improved the quality of visual features and boosted the performance of recent object discovery methods. However, the key challenge remains in extracting meaningful information from these features to discover objects without imposing restrictive assumptions about what constitutes an object.  DINOv1 \cite{caron_emerging_2021} produced attention maps that more faithfully reflect spatial object structure, making it better suited to our objective. Therefore, although DINOv2 \cite{oquab_dinov2_2023} offers state-of-the-art representation learning, after several tests, we opted for DINOv1 to preserve the spatial fidelity necessary for unsupervised object discovery (see Table~\ref{tab:experiments}).

This is mainly due to the spatial inconsistency introduced by the registers in DINOv2, which interfere with the interpretability and usefulness of attention maps and feature activations for downstream localization tasks. Since registers have no spatial correspondence, they tend to absorb attention in ways that degrade object saliency estimation and boundary precision. This problem was later addressed in Vision Transformers Need Registers \cite{darcet_vision_2023} by implementing a patch that mitigated the spatial inconsistency caused by registers. We compared our results using DINOv1, and DINOv2 with and without registers, as shown in Table~\ref{tab:experiments}. The iteration in which we introduced overlap (from version 0.6 to v0.8) proved to be critical in terms of accuracy. This is because objects with significant depth were being split across two different layers, and as a result, were either not discovered or detected as two separate objects. Introducing an overlap of 20–30\% led to a substantial improvement in performance. This parameter will be explored in future work to make it dynamic, rather than fixed as it is currently.

\vspace*{-5mm}
\begin{table}[]
\centering
\caption{Ablation study of the DADO framework on the VOC07 dataset.}
\label{tab:experiments}
\begin{tabularx}{\textwidth}{c|X|c}
\hline
\textbf{Version} & \textbf{Description} & \textbf{CorLOC} \\
\hline
v0.1 & Dinov2 & 53.74 \\
v0.1 & Dinov1 & 61.62 \\
v0.2 & Dinov1 and Depth & 69.64 \\
v0.4 & Dinov1 and Depth, with weights & 69.72 \\
v0.6 & Dinov1 and Depth-isolated layers & 72.70 \\
v0.8 & Dinov1 and Depth-isolated layers with 30\% overlap & 78.30 \\
v1.0 & Dinov1 and Depth-isolated, overlaped and dynamic bins & 78.32 \\
\hline
\end{tabularx}
\vspace*{-5mm}
\end{table}

\paragraph{\textbf{Discussions}}

\begin{figure}[t!]
    \centering
    \begin{subfigure}[t]{\textwidth}
        \centering
        \includegraphics[width=1\linewidth]{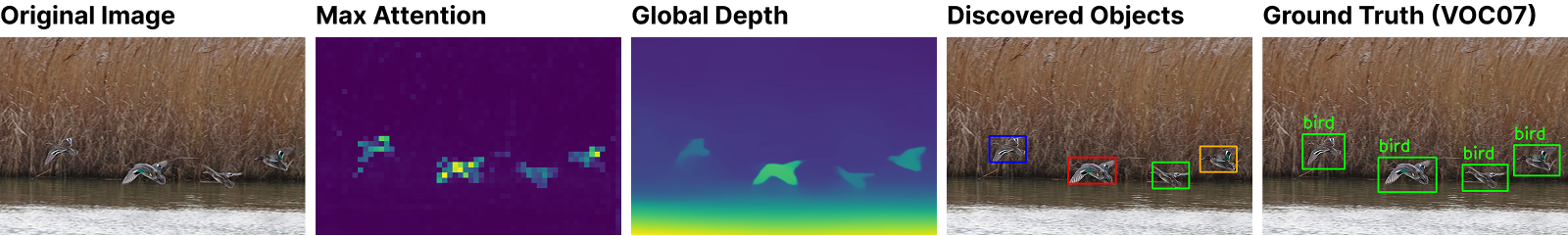}
        \caption{}
    \end{subfigure}%
    \hfill
    \begin{subfigure}[t]{\textwidth} 
        \centering
        \includegraphics[width=1\linewidth]{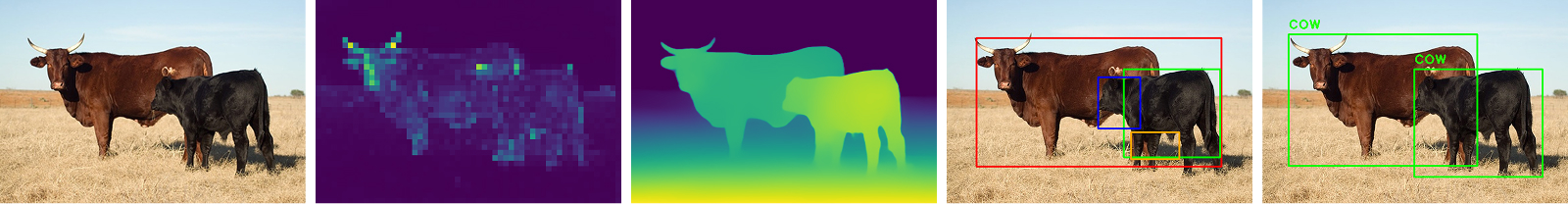}
        \caption{}
    \end{subfigure}
    \hfill
    \begin{subfigure}[t]{\textwidth}
        \centering
        \includegraphics[width=1\linewidth]{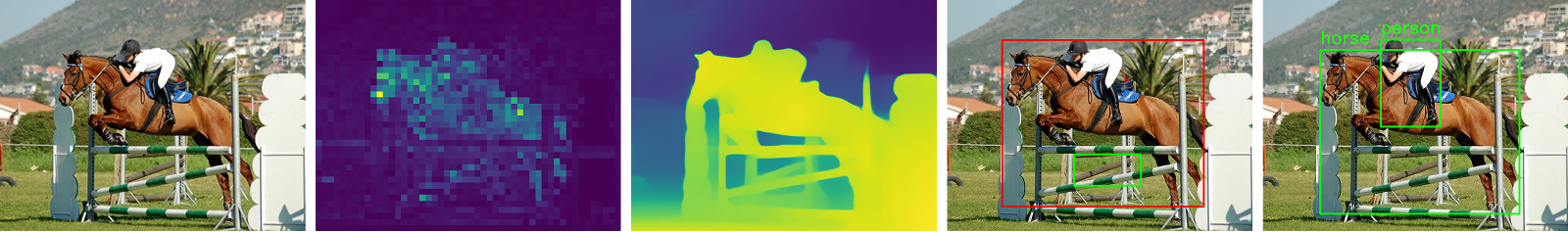}
        \caption{}
    \end{subfigure}
    \hfill
    \begin{subfigure}[t]{\textwidth}
        \centering
        \includegraphics[width=1\linewidth]{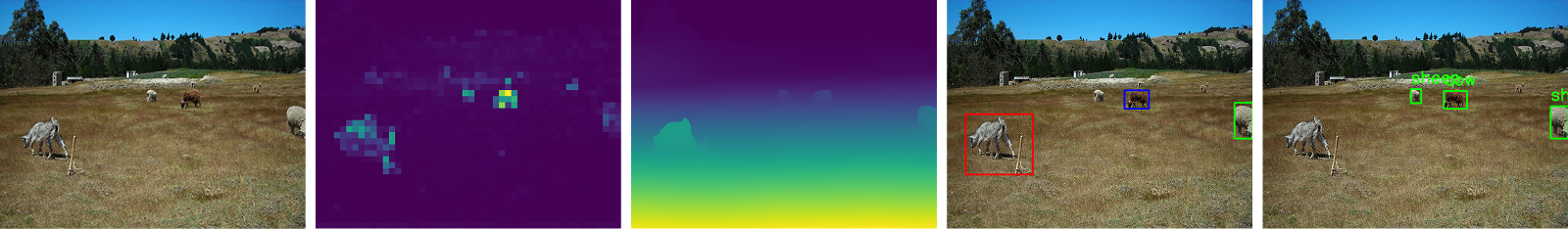}
        \caption{}
    \end{subfigure}
    \hfill
    \begin{subfigure}[t]{\textwidth}
        \centering
        \includegraphics[width=1\linewidth]{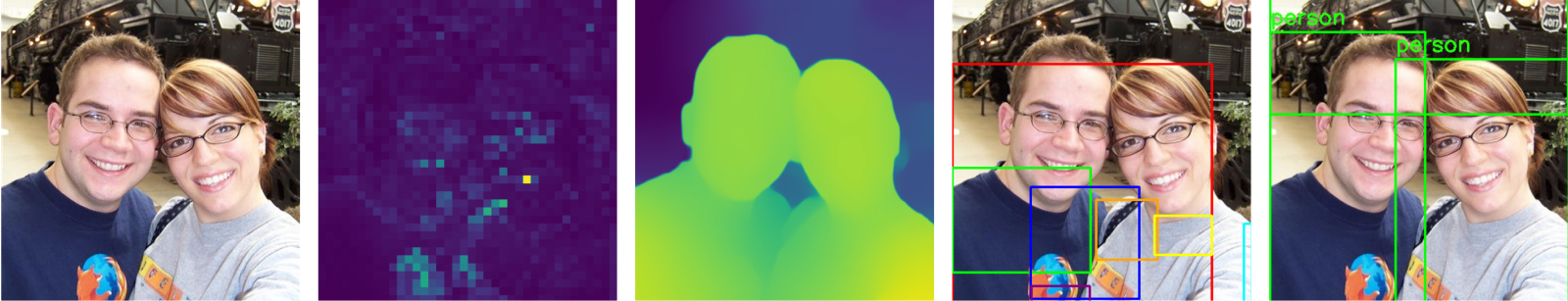}
        \caption{}
    \end{subfigure}
    \caption{Outputs of DADO. (a) Independent and isolated objects are effectively discovered by both attention mechanisms and depth cues. (b) Objects positioned in front of or behind others can be accurately separated using depth layers; in such cases, attention provides limited additional information. (c) Composite objects, such as the horse and rider, are very difficult to separate when they lie on the same plane—this represents the main weakness of our model. (d) Pascal VOC ground truth does not include the 'goat' class, but contains 'sheep'. DADO finds instances of both objects. (e) Separating two objects that are adjacent and on the same plane is particularly challenging for our model.}
    \label{fig:outputs}
    \vspace*{-5mm}
\end{figure}

One of the major challenges in object discovery is the absence of labels, which also implies a lack of semantics. Although the features obtained from models like DINO are of high quality, the lack of semantic information complicates the distinction of individual objects. They can appear in images as independent and isolated entities, as parts of larger composite objects, adjacent to others with no space in between, in front of or behind, leading to occlusions at different scales. Figure \ref{fig:outputs} shows examples of these situations and how DADO leverages attention and depth to resolve them. Note that our method successfully discovers both large and small objects, whether they are in the foreground or further in the background, and is also capable of separating them when they are overlapping (as in the case of the cows). However, it may also produce false positives in some situations, as shown in the fifth row.

Our approach, which incorporates depth layers, allows us to address some of these challenges, particularly occlusions. The core idea is that by leveraging depth information, we can achieve better separation of objects. Depth layers can help identify object boundaries and contours. This leads to more accurate object representation in the image, which ultimately improves feature quality and model performance.

Our findings reveal that object-centric images tend to produce attention maps of sufficient quality to detect unseen objects. In contrast, complex scenes containing important non-centric objects typically yield noisy and less discriminative attention maps. This noise arises because attention mechanisms  prioritize the most salient visual features, which in cluttered environments may belong to background textures or secondary objects. We observed that in such challenging cases, depth representations can significantly improve localization performance. Depth cues provide spatial information that helps distinguish foreground objects from the background, even when visual attention alone is ambiguous.
Separating the image into distinct depth planes and applying attention to each plane allows for the individualization of overlapping objects. 

By employing dynamic weights that adapt based on either attention features or depth layer representations, we further improve model quality. A higher number of objects in an image typically results in a more dispersed attention map, making object identification harder. Dynamic weights allow the model to adjust its focus on attention or depth.

These findings suggest that combining depth information with attention-based discovery methods could be a promising direction for improving unsupervised object discovery in complex real-world scenes.

\textit{\textbf{Limitations and Future Work}} Our method has difficulty  separating adjacent objects that lie in the same plane and lack visible gaps. Tokenization-based approaches, such as TokenCut and MOST, handle such cases more effectively. Additionally, low-level, non-semantic techniques, such as edge detection, superpixel grouping, and optimization-based methods such as graph cuts or watershed algorithms, can provide useful boundary cues without requiring semantic information. In future work, we aim to incorporate such strategies to improve performance in scenes with tightly packed objects.

\section{Conclusions}
\label{sec:conclusion}

In this paper, we introduce a novel unsupervised object discovery method in images based on inferred attention and depth features. Our model achieves state-of-the-art results and is highly adaptable to a wide range of visual scenes and object categories. These findings highlight the potential of combining mid-level cues for robust object discovery without supervision and pave the way for future extensions in image understanding and instance segmentation.

%
%
%
\bibliographystyle{splncs04}
\bibliography{object-discovery}

\end{document}